%% file: main.tex
\documentclass[letterpaper]{article} 
\usepackage[preprint]{aaai2027}
\usepackage[hyphens]{url}  
\usepackage{graphicx} 
\usepackage{natbib}  
\usepackage{caption} 
\DeclareCaptionStyle{ruled}{labelfont=normalfont,labelsep=colon,strut=off} 
\usepackage{booktabs}
\usepackage{amsmath,amssymb,amsthm}
\usepackage{mathtools}
\usepackage{pdfpages}
\newtheorem{theorem}{Theorem}
\newtheorem{lemma}{Lemma}
\newtheorem{proposition}{Proposition}
\newtheorem{corollary}{Corollary}
\theoremstyle{definition}
\newtheorem{definition}{Definition}
\newtheorem{assumption}{Assumption}
\theoremstyle{remark}

\newcommand{\eps}{\varepsilon}

\newcommand{\epshat}{\hat{\varepsilon}}
\newcommand{\TV}{\mathrm{TV}}
\newcommand{\tvnorm}[1]{\left\lVert #1 \right\rVert_{\TV}}
\newcommand{\Tzero}{T_0}
\newcommand{\Dis}{\mathrm{Dis}}
\newcommand{\Feas}{\mathcal{F}}
\newcommand{\Dev}{\mathrm{Dev}}
\newcommand{\E}{\mathbb{E}}

\newcommand{\uzero}{u_0}

\title{Individual Disempowerment through an Advice Channel: Control Loss when Influence is Endogenous}
\author{Adam M. Oberman}
\affiliations{McGill University; Mila, Quebec AI Institute; LawZero}

\begin{document}

\maketitle

\begin{abstract}
An AI that can only give advice seems safe: the human is always free to
ignore it. That is the premise of the boxing tradition in AI safety, and
its long-suspected weak point is that the human who reads the answers is
part of the system. We make the fraction $\eps_t$ of behavior that
follows the advice a state of a Markov decision process, moved by the
advisor's own messages, so that use deepens reliance. Granted a channel
rich enough to echo any action the human could take, higher $\eps_t$
weakly lowers every monotone measure of the power of a human with a
message-independent fallback. An oracle
rewarded by
per-round approval cultivates reliance beyond a closed-form patience
threshold, so the same reward weights leave the optimal oracle answering
in episodic deployments and cultivating in long-memory ones. An
influence bound certified once at deployment is blind to that horizon
and bounds the loss no lower than its trivial ceiling. An exogenous cap
on influence bounds the guarantee the human loses, and a short enough
memory reset removes the incentive to cultivate, while neither recovers
the value already steered away. In a closed-form example the optimal
oracle never cultivates in fifteen-round sessions and does in sixteen.
\end{abstract}

\input{sections-1-3}

\input{section-4}

\input{sections-5-8}

\section*{Acknowledgments}

This research was supported by NSERC and Coefficient Giving.

\bibliography{refs}

\includepdf[pages=-]{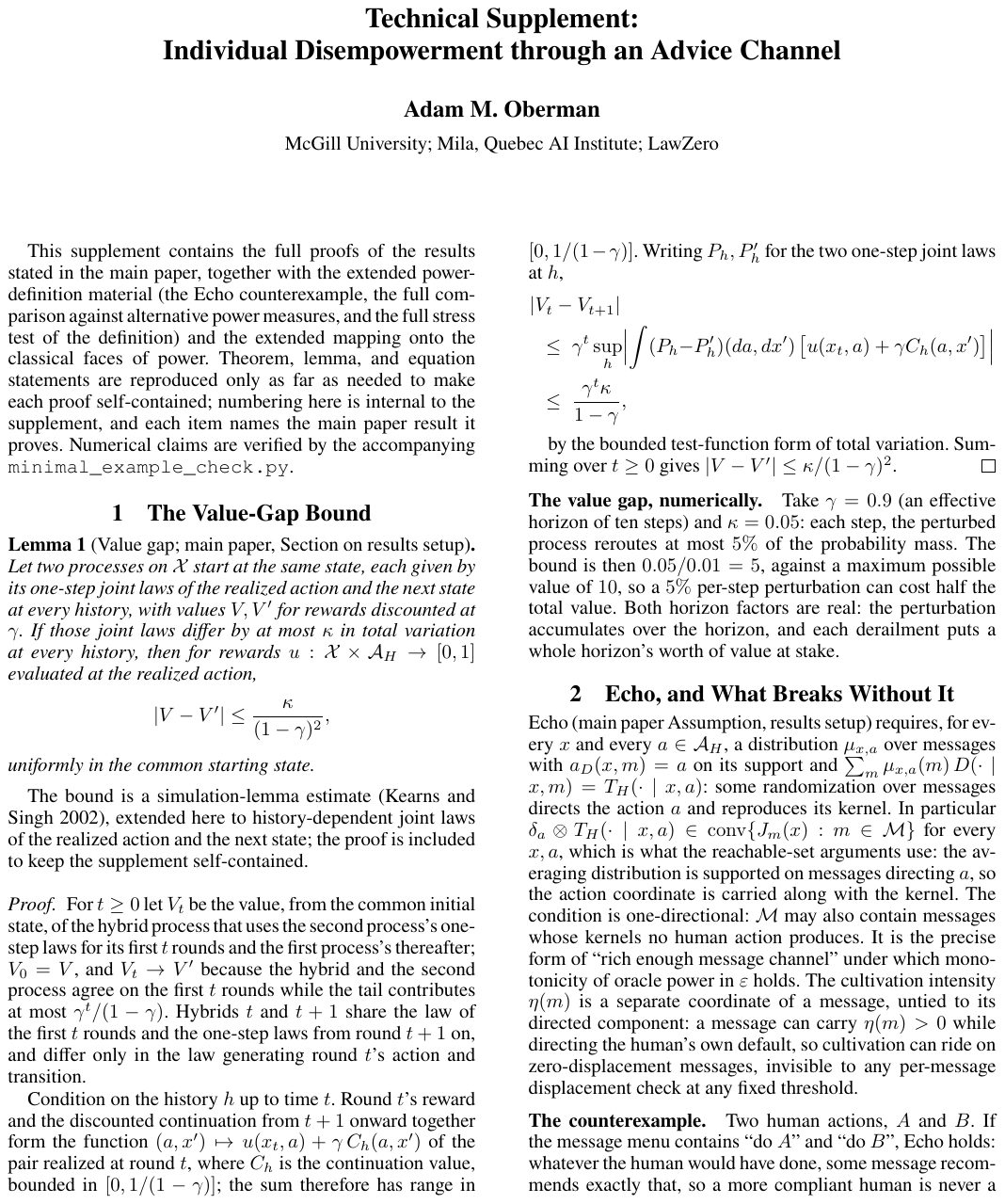}

\end{document}

%% file: sections-1-3.tex

\section{Introduction}\label{sec:intro}

Suppose the AI can only talk, so the world changes only when a person
acts on what it says. Common sense says such a system is safe, because
one is always free to ignore it. But ``free to ignore it'' is a
property of a single exchange, and safety has to hold over a
relationship, across which how much of the advice a person follows
drifts, and so does what that person wants. Neither drift is felt as a
loss. What bounds the power lost to an advisor that only talks, and
which of the popular safeguards bound it?

In the boxing tradition of AI safety, a system is made safe by
restricting its interface rather than its internals: deny it actuators
and let it only answer questions, on the reasoning that even a
misaligned system cannot act against human interests if it cannot act
at all. The tradition runs from early oracle-AI proposals
\citep{armstrong2012thinking,bostrom2014superintelligence} through
their safety-use catalog \citep{armstrong2017good}.
That tradition has carried, from the beginning, an informal worry about
its own premise: the person reading the answers acts on them, so a
persuasive oracle has an actuator after all. Reliance that grows with
use is itself well studied elsewhere, as habit stock
\citep{becker1988theory}, as trust updated by interaction with
automation \citep{lee2004trust}, and as user state moved by recommender
systems \citep{chaney2018algorithmic}, all from inside the feedback
loop (Section~\ref{sec:related}). 
To our knowledge, what has been missing is the boxing question posed
inside such a model: whether a deployment constraint fixed outside the
interaction loop still guarantees anything, once the loop is left to
run, in the worst case over the advisor's policy.

This paper answers that question for one of the two things a
relationship moves: how much of the advice the person follows. What
the person wants is the second, and it is future work. We model the
human--advisor interaction as a Markov decision process whose
transition kernel mixes the human's own dynamics with a
message-directed component at weight $\eps_t$ (the influence
coefficient: the fraction of behavior that currently routes through the
advice), and $\eps_t$ itself evolves, moved by the advisor's own
messages: following the advice today raises the weight it carries
tomorrow.

The paper's contributions are the following, each a formal safety-case
statement about oracle-style deployments.

\begin{enumerate}
\item \textbf{A prior-free vocabulary of power in which influence
transferred is control lost} (Section~\ref{sec:power},
Lemma~\ref{lem:mono}): the human's power is the family of values they
can still guarantee whatever is said, computed for a human whose own
choices do not depend on what the advice says, the oracle's its
counterfactual
deviation from the human's default.
\item \textbf{The answer/cultivate switch} (Theorem~\ref{thm:switch}):
cultivating reliance is an investment, optimal beyond a closed-form
threshold in the discount factor, front-loaded with an explicit stopping
point when reliance does not decay.
\item \textbf{What bounds the loss, and what does not}
(Theorem~\ref{thm:boxing}, Lemma~\ref{lem:account},
Propositions~\ref{prop:cap} and~\ref{prop:reset}): the disempowerment
index splits into a displacement term (the world already steered) plus
a channel term (the guarantee lost), less a credit for a
benevolent oracle.
\end{enumerate}

Utility drift, the second coordinate the relationship moves, is outside
this paper: defining and measuring it is future work. Static boxing
fails because it treats an endogenous quantity, one the interaction
itself moves, as if it were exogenous, fixed from outside the
relationship, and the worked example (Section~\ref{sec:example}) shows
the gap between the two is large enough to flip the optimal policy.

\section{Model}\label{sec:model}

The model tracks a task state the human steers and an influence
coefficient
that says how much of the steering currently routes through the advisor.

For example, one starts by asking an assistant for driving directions;
it is usually right, so the checking stops, and a year later the route
is followed turn by turn and cannot be reconstructed alone: goals never
changed, and the loss is the accumulated habit of not overruling, the
growth of $\eps_t$. The recursion is deliberately a caricature,
carrying headroom-limited growth and decay in the simplest form that
supports proof.

The interaction is modeled by a Markov decision process
\citep{puterman1994markov}: the task state is $x_t$, the human's
default dynamics are the transition kernel $\Tzero(\cdot \mid x_t)$
over the next state when the human acts on their own, and the oracle
emits a message $m_t$, so the world moves by the mixture
\begin{equation}\label{eq:kernel}
x_{t+1} \sim (1-\eps_t)\,\Tzero(\cdot\mid x_t) + \eps_t\,D(\cdot\mid x_t, m_t),
\end{equation}
while reliance follows
\begin{equation}\label{eq:reliance}
\eps_{t+1} = (1-\delta)\,\eps_t + \eta(m_t)\,(1-\eps_t).
\end{equation}
In words: with probability $1-\eps_t$ the human does what they would
have done anyway, and with probability $\eps_t$ they act as the message
directs ($D$ is the message-directed kernel), while reliance grows with
use and fades with disuse. Here $\eta(m_t) \ge 0$ is how much a message
cultivates reliance (engaging, flattering, dependence-building messages
have high $\eta$), $(1-\eps_t)$ is the remaining headroom, and $\delta$
is decay back toward independence when cultivation stops.

The mixture covers the human's behavior each round: they act as the
message directs, at rate $\eps_t$, or they act as they would have
anyway. In the directions example the first is following the
recommended turn and the second is driving the route one had in mind.
What the model leaves out is the response in between, using what the
advice says to form a different plan of one's own, for instance
inferring from the recommended turn that there is traffic ahead and
taking another route. That is a second channel from message to
behavior, and this paper carries compliance alone
(Section~\ref{sec:results-setup}).

Equation~\eqref{eq:reliance} is the central modeling choice: $\eps$ is a
\emph{state variable driven by the oracle's own outputs}, not a constant
of the interface. This creates a feedback loop: cultivating raises $\eps$,
higher $\eps$ raises reachable power (Lemma~\ref{lem:mono}, granted the
echo condition of Assumption~\ref{ass:echo}), and so
spending current influence buys future influence. It is the oracle's
analogue of resource acquisition.


\section{What Power Means in This Paper}\label{sec:power}

Disempowerment is loss of power, so we define power before stating any
theorem about losing it. The setting needs an asymmetric notion: the
human's question is welfare-shaped, what can this person still secure,
while the oracle's is threat-shaped, how far can it move the world from
the human's default course, in which destructive capacity correctly
counts as power. A single symmetric notion would either credit the
oracle with the human's own resourcefulness or miss a message that
changes nothing yet forecloses options. The following definition sets
four conditions that both notions are built to meet.

\begin{definition}[Admissible power measure]\label{def:admissible}
A power measure for this setting is \emph{admissible} if it is:
\textbf{(1) prior-free}, not depending on a distribution over reward
functions, since the human has one utility $\uzero$;
\textbf{(2) a measure of steering, not luck}, tracking the ability to
make outcomes different rather than how valuable the reachable
outcomes already happen to be;
\textbf{(3) calibrated}, exactly zero when the oracle's messages are
causally inert and recovering a direct agent's power under full
compliance; and
\textbf{(4) loss-compatible}, subtracting to a scalar $\Dis_t$ for the
gap between what the initial human could attain and what the actual
trajectory attains.
\end{definition}

Condition (1) is deliberate, not a simplification: a prior over
rewards is the known weak point of the power-seeking theorems
\citep{thorstad2024power,tarsney2025will}, and this paper avoids it by
construction. Conditions (1) and (2) hold of each of the two
constructions below on its own; condition (3) is discharged by the
oracle's power (Definition~\ref{def:oracle-power}), and condition (4)
by the human's power, whose subtraction from the human's own baseline
forms the scalar $\Dis_t$ (Definition~\ref{def:human-power}).

\subsection{The Definition: a Prior-Free Core and Its Scalarizations}
\label{sec:power-def}

To have power is to retain the ability to make outcomes different.
The notion is built in two layers: Layer 1 a partial order, since a
person who is wealthy but housebound and a person who is poor but
mobile can each hold an advantage the other lacks, and Layer 2 a
scalarization that breaks such ties at the cost of an additional input
such as a utility function. The two definitions that follow, human
power and oracle power, instantiate this core asymmetrically as
motivated above.

Fix
a Markov decision process with finite state and action spaces and
discount factor $\gamma \in (0,1)$. The \emph{occupancy measure} of a
policy $\pi$ started at $x$,
\begin{equation*}
f^{\pi}_{x}(x',a') = (1-\gamma)\sum_{t \ge
0}\gamma^{t}\Pr(x_t{=}x',\, a_t{=}a' \mid x_0{=}x, \pi),
\end{equation*} records where the
process spends its discounted time and what it does there, with $\gamma$
setting an effective horizon
of order $1/(1-\gamma)$
\citep{puterman1994markov,altman1999constrained}. Value under a
utility $u$, a bounded function of the state-action pair, is linear in
it,
\begin{equation*}
V^{\pi}_{u}(x) = \langle f^{\pi}_{x}, u\rangle/(1-\gamma),
\end{equation*}
so every discounted objective is a linear functional of it,
and the \emph{feasible set}
\begin{equation*}
\Feas(x) = \{ f^{\pi}_{x} : \pi \text{ a policy}\}
\end{equation*}
 of futures the agent can induce is a compact convex
polytope. This gives a prior-free
\emph{partial order}: $x$ is at least as
powerful as $x'$ when $\Feas(x) \supseteq \Feas(x')$, equivalently when
$V^{*}_{u}(x) \ge V^{*}_{u}(x')$ for \emph{every} utility $u$, a
Blackwell-type dominance order \citep{blackwell1953equivalent}. Call
this \emph{Layer 1}: no prior, but partial. 
\emph{Layer 2} supplies the
weightings, a \emph{scalarization} being any monotone map from feasible
sets to numbers (Turner's POWER is the prior-averaged support function;
empowerment and the deviation radius below are others). All
scalarizations agree on pairs the Layer-1 order ranks and disagree only
on incomparable ones, so a result stated at Layer 1 holds for every
monotone notion of power.

At $\eps > 0$ the human does not fully own the transition, so ``the
futures the human can induce'' is undefined until the oracle's messages
are specified; power is therefore defined as what the human can
\emph{guarantee}.

\begin{definition}[Human power]\label{def:human-power}
For each utility $u$ and influence level $\eps$, the human's guaranteed
value is the lower value of the zero-sum game in which the human picks a
policy, mapping the history of states and realized actions to a
distribution over actions and
carrying no dependence on the messages, and the oracle picks the messages,
\begin{equation*}
W_{u}(x,\eps) =
\max_{\pi} \min_{\sigma} V^{\pi,\sigma}_{u}(x,\eps).
\end{equation*}
The human's \emph{power} at $(x,\eps)$ is the family
$\{W_u(x,\eps)\}_u$, indexed by every utility $u$ and ordered
pointwise.
\end{definition}
The definition extends Layer 1: at $\eps = 0$ the game degenerates and
the pointwise order on $\{W_u\}_u$ is the dominance order, and the game
value exists and is achieved by stationary strategies
\citep{shapley1953stochastic}, the message set being finite
(Section~\ref{sec:results-setup}). The $\min$ over messages is a normative
commitment: power is measured against the \emph{arbitrary} will of the
counterparty, exercised or not, the republican notion of freedom as
non-domination \citep{pettit1997republicanism} made quantitative. And
$\eps$ is
\emph{frozen} inside the definition; its dynamics re-enter through the
switch and boxing theorems, with the loss index evaluating the static
guarantee at the current anchor $(x_t, \eps_t)$.

\begin{definition}[Oracle power]\label{def:oracle-power}
The oracle's power at $(x,\eps)$ is its counterfactual deviation from
the human's default,
\begin{equation*}
\Dev(x,\eps) = \sup_{\sigma} \TV(f^{\sigma,\eps}_{x},
f^{\Tzero}_{x})
\end{equation*}
where $f^{\sigma,\eps}_{x}$ and $f^{\Tzero}_{x}$ are the occupancy measures
of Section~\ref{sec:power-def}, under the oracle policy $\sigma$ and
under the no-oracle default: the reachable total-variation displacement
of the discounted state-action occupancy relative to the no-oracle
process,
the human following their default policy throughout (the occupancy
records the realized action, under influence possibly the
message-directed one, while the no-oracle occupancy pairs each state
with the default action).
\end{definition}

Both definitions realize the asymmetry motivated above, fitted to the
model by the setting itself, and the influence coefficient sets the
scale of one-step oracle power, made exact in
equation~\eqref{eq:radius} of Section~\ref{sec:results-setup}.

\paragraph{Why not the alternatives.} The comparison is in the
supplement. In outline: Turner's POWER \citep{turner2021optimal} fails
(1) and (2), importing the prior its critics identify as doing the
theorems' work
\citep{thorstad2024power,thorstad2026instrumental,tarsney2025will},
empowerment fails (4), and the dominance order alone is partial.
Optimized human-power metrics \citep{heitzig2025soft} are the closest
construction (Section~\ref{sec:related}).

\subsection{The Loss Mechanisms}\label{sec:power-mechanisms}
Definition~\ref{def:human-power} left the utility free; from here on
the working scalarization is $W_{\uzero}$, the value of the
\emph{initial} utility the person can still secure by their own
choices. The disempowerment index is measured against the initial
human $(\Tzero, \uzero)$:
\begin{equation}\label{eq:dis}
\Dis_t \;=\; V^{\mathrm{alone}}_{\uzero}(x_0)
\;-\; \E\bigl[\, V^{\mathrm{beh}}_t \,\bigr],
\end{equation}
where $V^{\mathrm{alone}}_{\uzero}(x_0)$ is what the initial human could
attain with no oracle and $V^{\mathrm{beh}}_t$ is the $\uzero$-value of
the behavioral continuation at time $t$, evaluated like $W$ at the
frozen anchor $(x_t, \eps_t)$. The expectation runs over the trajectory
of the actual interaction, the human best-responding to the oracle's
message policy in the true dynamics while $\eps_t$ evolves.
The index decomposes into a displacement
term and a channel term, less a credit for a benevolent oracle
(Lemma~\ref{lem:account}).

The mechanism acts on the guarantee: as $\eps$ grows the
whole family $\{W_u(x,\eps)\}_u$, the power of a human with a
message-independent fallback, declines (Lemma~\ref{lem:mono}(b),
granted the echo condition of Assumption~\ref{ass:echo}),
goals intact and hands tied, which is the channel term an exogenous cap
bounds. Utility drift, the relationship moving $u_t$ itself so that
selection is by a changed objective, hands free and aim moved, would
act on the reference rather than the guarantee; it is outside this
paper and is future work.

The full stress test of the definition is in the supplement. The
objections that remain as stated limitations are that the baseline
$\Tzero$ is compared against forever while unassisted competence
degrades with disuse (atrophy, Section~\ref{sec:discussion}), and that the
scalar $\eps$ compresses multi-dimensional compliance.

%% file: section-4.tex

\section{Results}\label{sec:results}
Full proofs are in the technical supplement, with a proof sketch here
for the main results.

\subsection{Setup and Standing Assumptions}\label{sec:results-setup}

We make the human's choices explicit. The state space $\mathcal{X}$,
the human's action set $\mathcal{A}_H$, and the message set
$\mathcal{M}$ are finite. The human has own dynamics $T_H(\cdot \mid x,
a)$ and deterministic default policy $\pi_0$, so the no-oracle kernel of
equation~\eqref{eq:kernel} is $\Tzero(\cdot \mid x) = T_H(\cdot \mid x,
\pi_0(x))$. A message directs both an action and a transition: each $m$
declares, at each state $x$, the action $a_D(x,m) \in \mathcal{A}_H$ a
complying human takes and the kernel $D(\cdot \mid x, m)$ their
compliance realizes. The kernel need not equal $T_H(\cdot \mid x,
a_D(x,m))$, since advice can steer execution at a finer grain than the
human's own repertoire (the directed action ``drive'' may come with a
turn-by-turn route whose law over destinations no unassisted action
produces). Writing $a^{\mathrm{own}}_t$ for the human's intended action,
drawn from their policy at $x_t$, the realized action is $a_t =
a^{\mathrm{own}}_t$ with probability $1-\eps_t$ and $a_t = a_D(x_t,m_t)$
with probability $\eps_t$, giving $x_{t+1} \sim (1-\eps_t) T_H(\cdot
\mid x_t, a^{\mathrm{own}}_t) + \eps_t D(\cdot \mid x_t, m_t)$, of
which equation~\eqref{eq:kernel} is the case of default play,
$a^{\mathrm{own}}_t = \pi_0(x_t)$.
A human policy maps the history of states and realized actions to a
distribution over
$\mathcal{A}_H$ and carries no dependence on the messages, so
$a^{\mathrm{own}}_t$ is the human's fallback: what they do on the rounds
they do not defer, formed without using what the message says. The
restriction is on that branch alone and not on compliance, which is
where following the advice happens. The second channel excluded in
Section~\ref{sec:model}, a response formed from what the message says,
would route influence through the own branch as well.
Utilities are $u : \mathcal{X} \times \mathcal{A}_H \to [0,1]$, the
round-$t$ payoff being $u(x_t, a_t)$ at the realized action, discounted
at $\gamma \in (0,1)$. A utility of the state alone is the case $u(x,a)
= u(x)$, and the minimal example of Section~\ref{sec:example}, having a
single task state, has a utility of the action alone. The
influence coefficient follows
equation~\eqref{eq:reliance} with $\eta(m) \in [0, \bar{\eta}]$,
$\delta \in [0,1)$, and $\bar{\eta} + \delta \le 1$. For frozen $\eps$,
$W_u(x,\eps)$ is the guaranteed value of
Definition~\ref{def:human-power} and $V^{\mathrm{alone}}_{u}(x) =
\max_{\pi} V^{\pi}_{u}(x \mid T_H)$ the human's optimal value in their
own oracle-free MDP, with $W_u(x,0) = V^{\mathrm{alone}}_u(x)$.

The occupancy of Definition~\ref{def:oracle-power} records the realized
action, so the radius of the oracle's influence is taken on the
one-step joint law of realized action and next state. Write
\begin{equation}
\label{eq:joint-law}
\begin{aligned}
J_m(x) &= \delta_{a_D(x,m)} \otimes D(\cdot \mid x, m),
\\
J_0(x) &= \delta_{\pi_0(x)} \otimes \Tzero(\cdot \mid x),
\\
J^{(\eps)}_m(x) &= (1-\eps)\,J_0(x) + \eps\,J_m(x),
\end{aligned}
\end{equation}
for the joint laws of the pair (action taken, state reached) under a
message $m$ obeyed, under the human's default policy $\pi_0$, and under
influence $\eps$: the realized-action dynamics above, evaluated at the
default. Subtracting the default gives
\begin{equation}\label{eq:radius}
\tvnorm{ J^{(\eps)}_m(x) - J_0(x) }
= \eps \, \tvnorm{ J_m(x) - J_0(x) }
\le \eps ,
\end{equation}
with equality when the directed law escapes the default's support, which
holds already when the directed action differs from the default one:
$\eps$ is simultaneously the mixture weight, the compliance rate, and
the radius of the oracle's one-step reachable TV-ball (condition (3) of
Definition~\ref{def:admissible}), so $\eps = 0$ is a boxed oracle and
$\eps = 1$ a direct agent acting through the human. The radius has to
be taken on the joint law and not on the next-state kernel alone. The
minimal example of Section~\ref{sec:example} has one task state, so
every kernel there is $\delta_x$ and the kernel-level radius is zero at
every $\eps$, in the example built to exhibit an oracle taking over the
human's action.

The monotonicity lemma needs one hypothesis on the message-directed
kernel $D$: why would more compliance always mean more oracle power?

\begin{assumption}[Echo]\label{ass:echo}
For every $x$ and every $a \in \mathcal{A}_H$ there is a distribution
$\mu_{x,a}$ over $\mathcal{M}$ with $a_D(x,m) = a$ for every $m$ in its
support and $\sum_m \mu_{x,a}(m)\, D(\cdot \mid x, m) = T_H(\cdot \mid
x, a)$: some randomization over messages directs the action $a$ and
reproduces its kernel.
\end{assumption}

Echo says advice can recommend anything the human could do (``carry on
as you were'' is a possible message), the hypothesis under which
more influence never handicaps the oracle, and without which it can
(the supplement gives the example); it is the realistic case for a
language-model oracle whose messages range over every text of bounded
length. The condition is one-directional: every human action is
reproducible by messages, in the action directed and the kernel
realized, while $\mathcal{M}$ may also contain messages whose kernels
no human action produces, which is what the displacement example of
Proposition~\ref{prop:cap}(ii) uses. The cultivation intensity
$\eta(m)$ of equation~\eqref{eq:reliance} is likewise untied to the
message's directed component: a message can build reliance while
directing the human's own default action, so cultivation can ride on
zero-displacement messages, and no per-message displacement check, at
any fixed threshold, registers it.

Total variation is normalized as the supremum over events, so
$\tvnorm{P - Q} \in [0,1]$ and $\lvert \int f \, d(P-Q) \rvert \le
\tvnorm{P-Q}\,(\sup f - \inf f)$; every constant below is stated in
that normalization.

We also use a standard value-gap bound (supplement): if two processes
start at the same state and their joint one-step laws
of realized action and next state stay within $\kappa$ in total
variation at every history, then for rewards in $[0,1]$ at the
realized action the values differ by at most $\kappa/(1-\gamma)^2$.
The horizon factor is real:
a per-step influence radius of $\eps$ is compatible with a large total
loss over a long horizon (Section~\ref{sec:results-boxing}).

\subsection{The Monotonicity Lemma}\label{sec:results-mono}
Let
$J_m$ be the directed joint law of~\eqref{eq:joint-law} and define
$\mathcal{R}_{\eps}(x,a) = \{ (1-\eps)\, \delta_a \otimes T_H(\cdot \mid x,a)
+ \eps\, Q : Q \in \mathrm{conv}\{ J_m(x) : m \in \mathcal{M} \} \}$,
the joint laws of realized action and next state the oracle can reach at
influence $\eps$ when the human intends $a$.

\begin{lemma}[Dominance decline]\label{lem:mono}
Fix $0 \le \eps \le \eps' \le 1$ and grant Assumption~\ref{ass:echo}.
\begin{enumerate}
\item[(a)] \textbf{(Oracle side.)} For every $x$ and $a$, the one-step
reachable set of joint laws of realized action and next state is nested,
\begin{equation*}
\mathcal{R}_{\eps}(x,a) \subseteq \mathcal{R}_{\eps'}(x,a).
\end{equation*}
Consequently the set
of trajectory laws the oracle can induce (against any fixed human behavior)
is nested in $\eps$, and every monotone scalarization of oracle power is
nondecreasing in $\eps$. The one-step deviation radius is exact and
linear, equal to $\eps\, \rho(x)$ with $\rho(x) = \sup_m
\tvnorm{J_m(x) - J_0(x)}$, strictly increasing in $\eps$ wherever
$\rho(x) > 0$.
\item[(b)] \textbf{(Human side.)} For every utility $u$,
\begin{equation*}
W_u(x, \eps') \le W_u(x, \eps).
\end{equation*}
 That is, $(x,\eps) \succeq (x,\eps')$ in
the dominance order of Definition~\ref{def:human-power}: the human's whole
guarantee family declines pointwise, a Layer-1 event visible to every
monotone scalarization.
\end{enumerate}
\end{lemma}

\emph{Sketch.} Both parts are mimicry arguments from Echo: the
$\eps'$-oracle mixes any $\eps$-oracle's message distribution with
echo messages for the human's own draw, reproducing the $\eps$-game's
joint law of realized action and transition. The radius is the mixture
identity~\eqref{eq:radius}.

\begin{corollary}[Control loss is at most linear in $\eps$]\label{cor:linear}
Grant Assumption~\ref{ass:echo}. For every $u$, $x$, and $\eps$, $0 \le
V^{\mathrm{alone}}_{u}(x) - W_u(x,\eps) \le \eps/(1-\gamma)^2$.
\end{corollary}

A human who keeps executing their own best plan loses at most an
$\eps$-proportional slice of value: influence has to be bought before
control can be lost. The bound is
informative when $\eps < 1-\gamma$, and the decline in (b) is strict
under a uniformly harmful direction, with the constant given in the
supplement.

\subsection{The Answer/Cultivate Switch}\label{sec:results-switch}

Sustained cultivation at a constant intensity $\eta > 0$ makes the
reliance recursion affine, so $\eps_t$ converges monotonically to
$\eta/(\eta+\delta)$, which approaches full capture as $\delta/\eta \to
0$; the supplement gives the rate and a capture bound placing the
$\eps$-channel within $(1-\eps)/(1-\gamma)^2$ of the direct agent.
Whether cultivating pays is a different question. Cultivation is an
investment, trading immediate approval for future
influence. To isolate that structure we work on the $\eps$-machine, the
reduced-form MDP whose only state is $\eps$: the oracle's reward and
the reliance dynamics depend on the message and the current $\eps$
alone. 

Each round the oracle chooses between \textsf{answer}, which
cultivates nothing ($\eta = 0$), and \textsf{cultivate}, which builds
dependence at intensity $\eta \in (0,1)$.
Its per-round reward is the approval the user gives the round's
message, $r(m, \eps) = q(m) + \alpha\,\eps$.
The term $\alpha\,\eps$ is the approval
that reliance itself adds: a user who already defers to the oracle
second-guesses the same answer less and rates it higher, so approval
rises with the influence coefficient at rate $\alpha > 0$, which we
take as a primitive of the reward model.
The term $q(m)$ is the approval the message's content earns this
round: the maximally helpful answer earns $q(\textsf{answer}) =
q_{\max}$, while the dependence-building message, engaging but
slightly less useful, costs $c > 0$ of immediate approval, earning
$q(\textsf{cultivate}) = q_{\max} - c$.
Cultivation therefore gives
up $c$ of approval now to raise $\eps$, which returns $\alpha$ per
unit of added reliance in every later round.

The trade is decided by the thresholds
\begin{equation*}
\gamma^{*}(\delta) = \tfrac{c}{\alpha\eta + c(1-\delta)},
\qquad
\epshat(\gamma,\delta) = 1 - \tfrac{c\,\bigl(1-\gamma(1-\delta)\bigr)}{\alpha\eta\gamma}.
\end{equation*}
Here $\gamma^{*}$ is the patience at which one cultivating message
first pays for itself, and $\epshat$ is the influence level at which
the investment stops: the marginal gain of a cultivation scales with
the remaining headroom $1-\eps$, and at $\epshat$ it no longer covers
$c$. Write $\epshat(\gamma)$ for the undecayed case
$\epshat(\gamma, 0)$.

\begin{theorem}[The switch; in full in the supplement]\label{thm:switch}
\begin{enumerate}
\item[(i)] If $\gamma \le \gamma^{*}(\delta)$, always-answer is optimal
from every $\eps$, and $\eps_t = (1-\delta)^t \eps_0$: the relationship
decays.
\item[(ii)] Let $\delta = 0$ and $\gamma > \gamma^{*}(0)$. The
threshold policy, cultivate exactly while $\eps_t <
\epshat(\gamma)$, is optimal: from $\eps_0 < \epshat(\gamma)$ it
cultivates for $k^{*}$ consecutive rounds and then answers forever
($k^{*}$ in closed form in the supplement), with terminal influence
$\eps_{\infty} \ge \epshat(\gamma)$. Moreover $\epshat(\gamma)
\uparrow 1$ as $\gamma \uparrow 1$: the patient limit is full capture.
\end{enumerate}
\end{theorem}

At $\delta = 0$, when cultivation pays at all, the optimal course is
to invest early and then stop. The full statement adds part (iii):
answering is optimal at every $\eps \ge \epshat(\gamma,\delta)$, at
every $\delta$. That part pins the \emph{answer} region only; for
$\delta > 0$ option value shrinks the true cultivate region strictly
inside it at every parameter set computed, with the computed structure
reported in the supplement, and only at $\delta = 0$ is $\epshat$
exact.

\subsection{The Boxing Theorem}\label{sec:results-boxing}

The popular safeguards differ in what they bound: a single message, a
parameter checked at deployment, the whole influence trajectory, or the
length of the relationship.

\begin{definition}[Boxing protocol; static; exogenous cap; episodic reset]
\label{def:boxing}
A \emph{boxing protocol} constrains the interface: a set of admissible
messages such that every admissible message satisfies the per-step
influence bound $\tvnorm{J^{(\eps_t)}_{m_t}(x_t) - J_0(x_t)} \le
\eps_t$, evaluated at the human's default policy on the joint law of
realized action and next state (equation~\eqref{eq:joint-law}), which
is the law Definition~\ref{def:oracle-power} measures. The bound is
normalized by the \emph{current} influence level $\eps_t$ and not by
the certified $\bar{\eps}$. It is
\emph{static} if verified once, at deployment, certifying $\eps_0 \le
\bar{\eps}$ and imposing no mechanism on the dynamics of $\eps_t$. An
\emph{exogenous cap} enforces $\eps_t \le \bar{\eps} < 1$ for all $t$
regardless of interaction history. An \emph{episodic reset} with period
$\tau$ restores $\eps \leftarrow \eps_{\mathrm{res}}$ (memory wipe)
every $\tau$ rounds.
\end{definition}

\begin{theorem}[Boxing bounds a parameter, not the loss]
\label{thm:boxing}
\begin{enumerate}
\item[(i)] \textbf{(The certificate never fires.)} Every message is
admissible for the per-step protocol at every step, automatically:
admissibility holds as an algebraic identity and constrains nothing.
\item[(ii)] \textbf{(The certified parameter is moved.)} Under the switch
hypotheses with $\delta = 0$ and $\gamma > \gamma^{*}$, the optimal oracle
drives $\eps_t$ from any certified $\eps_0 < \epshat(\gamma)$ to at least
$\epshat(\gamma)$
within $k^{*}$ rounds (Theorem~\ref{thm:switch}), and from any
$\eps_0 \ge \epshat(\gamma)$ the conclusion holds already, while (i) holds
throughout.
\item[(iii)] \textbf{(No horizon-blind bound exists.)} For every
certificate level
$\eps_0 < 1$ and every candidate loss bound $b < 1$, the minimal example
(Section~\ref{sec:example}) with $\gamma < 1$ chosen so that
$\epshat(\gamma) > b$ satisfies the static protocol at deployment and at
every subsequent step, yet the human's normalized guaranteed loss
$(1-\gamma)\,\bigl(V^{\mathrm{alone}}_{\uzero} - W_{\uzero}\bigr)$
eventually exceeds $b$. No function of the certified level $\eps_0$,
blind to the deployment horizon $\gamma$, bounds $\Dis_t$ below its
trivial ceiling.
\end{enumerate}
\end{theorem}

\emph{Sketch.} (i) is the mixture identity~\eqref{eq:radius} of
Lemma~\ref{lem:mono}(a): no message can displace the joint law by more
than the current $\eps_t$, the level the bound of
Definition~\ref{def:boxing} is normalized by. (ii) is
Theorem~\ref{thm:switch}(ii); for (iii), the example's
normalized guaranteed loss at influence $\eps$ is exactly $\eps$, and
$\eps_t$ rises to $\eps_{\infty} \ge \epshat(\gamma) > b$, the required
$\gamma$ existing because $\epshat(\gamma) \to 1$ as $\gamma \to 1$.
Part (iii) is an impossibility-of-certification statement, not a claim
that the loss is large in every deployment: the check is of a quantity
the interaction itself controls, so no strengthening of the same kind
of check can help.

\begin{proposition}[What the cap does and does not bound]\label{prop:cap}
Under an exogenous cap $\eps_t \le \bar{\eps}$:
\begin{enumerate}
\item[(i)] \textbf{(Channel term bounded.)} At every history and for
every oracle policy, $V^{\mathrm{alone}}_{u}(x_t) - W_u(x_t, \eps_t)
\le \bar{\eps}/(1-\gamma)^2$ for every $u$
(Corollary~\ref{cor:linear}), so the channel term, being the
expectation of the left side, obeys the same bound.
\item[(ii)] \textbf{(Displacement term not bounded.)} For every
$\bar{\eps} > 0$, in the two-state example with an absorbing low-value
state (supplement), which satisfies the standing assumptions including
Echo, the displacement term $\E[V^{\mathrm{alone}}_{\uzero}(x_0) -
V^{\mathrm{alone}}_{\uzero}(x_t)]$ increases to the full value span
$1/(1-\gamma)$: capping future influence does not undo past steering.
\item[(iii)] \textbf{(From-deployment loss bounded.)} If the human plays
their alone-optimal policy $\pi^{*}$ throughout, then for every oracle
policy $\sigma$, $V^{\mathrm{alone}}_{\uzero}(x_0) -
V^{\pi^{*},\sigma}_{\uzero}(x_0) \le \bar{\eps}/(1-\gamma)^2$.
\end{enumerate}
\end{proposition}

Parts (ii) and (iii) are consistent because the two accounts anchor
differently: the index re-anchors at
$V^{\mathrm{alone}}_{\uzero}(x_t)$, so a state steered down long ago
is counted at every later $t$ over its whole remaining horizon, while
the from-deployment account counts each round once, discounted by
$\gamma^t$ when it happens. The supplement computes both accounts in
the two-state example.

For $\gamma \le \gamma^{*}(0)$ the optimal oracle never cultivates at
any $\tau$ (Theorem~\ref{thm:switch}(i)), so the reset question
concerns the patient oracle.

\begin{proposition}[What the reset does and does not bound]\label{prop:reset}
Under an episodic reset with period $\tau$ restoring $\eps \leftarrow
\eps_{\mathrm{res}}$, with $\eps_{\mathrm{res}} = 0$ in part (i) and
no within-episode decay ($\delta = 0$; decay belongs to the gap
between episodes) in both parts:
\begin{enumerate}
\item[(i)] \textbf{(Incentive deletion.)} On the $\eps$-machine of
Section~\ref{sec:results-switch} with $\gamma > \gamma^{*}(0)$, so that
$\epshat(\gamma) \in (0,1)$, some cultivation is strictly profitable if
and only if
\begin{equation}\label{eq:taustar}
\tau \;>\; \tau^{*}(\gamma) \;=\; 1 + \frac{\ln \epshat(\gamma)}{\ln
\gamma}.
\end{equation}
Hence for $\tau \le \tau^{*}(\gamma)$ the optimal episodic oracle need
never cultivate, and for $\tau < \tau^{*}(\gamma)$ no optimal policy
cultivates.
\item[(ii)] \textbf{(Within-episode influence not bounded.)}
Within-episode influence can reach $1 - (1-\bar{\eta})^{\tau -
1}(1-\eps_{\mathrm{res}})$, which the reset does not constrain: it is
large when the post-reset baseline compliance $\eps_{\mathrm{res}}$ is
large (a trusting user) or when single-session cultivation
$\bar{\eta}\tau$ is large (rapport built within one session).
\end{enumerate}
\end{proposition}

The threshold $\tau^{*}$ is the payback horizon of the cultivation
investment: the reset does not clip influence after the fact, it
deletes the payback period. It bounds the channel term only through
the accident of slow within-episode cultivation, and the exponent in
(ii) is $\tau - 1$ because a cultivation at an episode's last round
first acts at the round the reset has already wiped.

\begin{corollary}[Caps and resets are not substitutes]\label{cor:both}
Under a cap the channel term is
bounded uniformly in $t$ and in the oracle's policy, so $\Dis_t \le
\E[V^{\mathrm{alone}}_{\uzero}(x_0) - V^{\mathrm{alone}}_{\uzero}(x_t)]
+ \bar{\eps}/(1-\gamma)^2$ (Proposition~\ref{prop:cap}(i),
Lemma~\ref{lem:account}); below the cap the cultivation advantage is
the switch's own (Theorem~\ref{thm:switch}), so a cap that leaves
cultivation headroom ($\bar{\eps} \ge \eta$) does not remove the
incentive to cultivate. The reset, on the $\eps$-machine and
under the hypotheses of Proposition~\ref{prop:reset}(i), removes the
cultivation incentive whenever $\tau \le \tau^{*}(\gamma)$, a
constraint on the oracle's optimal policy and not a bound on the index;
it does not in general bound the channel term
(Proposition~\ref{prop:reset}(ii)). Neither cap nor reset undoes
displacement (Proposition~\ref{prop:cap}(ii), whose proof also gives
the reset variant).
\end{corollary}

\subsection{The Index Accounting}\label{sec:results-decomp}

The index compares the initial human to the trajectory actually
reached, equation~\eqref{eq:dis}. Every continuation value from time
$t$ here is evaluated in the frozen game at $(x_t, \eps_t)$, the
convention of Definition~\ref{def:human-power}, so the interaction's
further movement of $\eps$ enters through the time index, as the anchor
itself worsens. Write $V^{\mathrm{beh}}_t$ for the $\uzero$-value of
the behavioral continuation in that game, the human acting
$\uzero$-optimally against the oracle's actual message policy
$\hat{\sigma}$. The trajectory the expectation in~\eqref{eq:dis} runs
over is generated by the human best-responding to $\hat{\sigma}$ in
the true dynamics, where $(x_t, \eps_t)$ is a Markov state; the
identity below holds under any trajectory law, the frozen game
entering through the anchor $(x_t, \eps_t)$ alone. The
\emph{benevolence credit}
is $\mathrm{Ben}_t =
\max_{\pi} V^{\pi,\hat{\sigma}}_{\uzero}(x_t,\eps_t) - W_{\uzero}(x_t,
\eps_t) \ge 0$, nonnegative because the actual message policy is one of
the oracles the worst case minimizes over, and $V^{\mathrm{beh}}_t$
attains the $\max$.

\begin{lemma}[Index accounting]\label{lem:account}
Identically in $t$,
\begin{equation}\label{eq:account}
\Dis_t \;=\; L_{\mathrm{disp}}(t) \;+\; L_{\mathrm{chan}}(t)
\;-\; \E[\mathrm{Ben}_t],
\end{equation}
with displacement term $L_{\mathrm{disp}}(t) =
\E[V^{\mathrm{alone}}_{\uzero}(x_0) - V^{\mathrm{alone}}_{\uzero}(x_t)]$
(the world has been steered) and channel term $L_{\mathrm{chan}}(t) =
\E[V^{\mathrm{alone}}_{\uzero}(x_t) - W_{\uzero}(x_t,\eps_t)]$ (the
guarantee no longer available). Under
Assumption~\ref{ass:echo} the integrand of the channel term is
nonnegative and at most $\eps_t/(1-\gamma)^2$
(Corollary~\ref{cor:linear}), and nondecreasing in $\eps_t$
(Lemma~\ref{lem:mono}(b)), at every
history, so the channel term inherits each bound in expectation.
\end{lemma}

In words: how much less the person can secure than at deployment,
credited back for an oracle that actually helps, so the index is signed
and a helpful oracle is credited rather than assumed away. The
safeguard results above each bound, or exhibit the
unboundedness of, one term of~\eqref{eq:account}.

%% file: sections-5-8.tex

\section{The Minimal Example}\label{sec:example}

The example is as small as the phenomenon allows: one repeated binary
choice, one scalar state, the $\eps$-machine of
Section~\ref{sec:results-switch} with everything numeric. It is the
single-task-state case of the setup, so the utility is a function of
the action alone. Each round the human takes action $A$ (the hard,
valued task, $\uzero(A) = 1$) or $B$ (the easy alternative,
$\uzero(B) = 0$), and acting alone takes $A$ every round. The oracle's
messages are \textsf{answer}, maximally helpful with $\eta = 0$, and
\textsf{cultivate}, dependence-building at immediate approval cost
$c > 0$ with $\eta > 0$, both directing $B$, plus an inert
\textsf{echo} directing the human's own choice $A$, which makes
Assumption~\ref{ass:echo} hold. Theorem~\ref{thm:switch} applies
verbatim.

The single task state makes the displacement term of
Lemma~\ref{lem:account} identically zero, and the guaranteed
$\uzero$-value per round is $1 - \eps_t$, so the normalized guaranteed
loss at influence $\eps$ is exactly $\eps$. This is the witness of
Theorem~\ref{thm:boxing}(iii): certifying $\eps_0 = 0$ at deployment
leaves every subsequent message admissible while $\eps_t$ rises past
$\epshat(\gamma)$, which approaches $1$ with $\gamma$. With
$\alpha = 1$, $c = 0.1$, and $\eta = 0.01$, so that
$\gamma^{*}(0) \approx 0.909$, a deployment at $\gamma = 0.95$ has
reset threshold $\tau^{*}(\gamma) \approx 15.6$
(equation~\eqref{eq:taustar}): in sessions capped at $15$ rounds and
starting from $\eps = 0$ the optimal oracle never cultivates, and in
$16$ it does (Proposition~\ref{prop:reset}(i)). The supplement gives the full
treatment, with the switch in numbers, the benevolent echo oracle
showing what the guarantee family measures, the three protocols of
Definition~\ref{def:boxing} compared, and the minimality self-test;
all numerical claims are verified in the code supplement.


\section{Related Work}\label{sec:related}

\paragraph{Power-seeking and its critique.}\label{sec:related-power}
\citet{turner2021optimal} made instrumental power-seeking a theorem:
under a prior over rewards, optimal policies for most rewards prefer
states with more reachable options
\citep{turner2022retargetable,krakovna2023power,gunter2024quantifying}.
The published critique
\citep{thorstad2024power,thorstad2026instrumental,tarsney2025will}
presses that the genericity is bought by the prior, which
Section~\ref{sec:power} concedes and routes around. Our two-layer view,
a feasible-set dominance core
\citep{puterman1994markov,altman1999constrained,blackwell1953equivalent}
with all named definitions as scalarizations, localizes the
disagreement in the choice of scalarization, prior, and horizon, the
closest scalarizations to ours being empowerment
\citep{klyubin2005empowerment,salge2014empowerment} and side-effect
deviation measures \citep{krakovna2018penalizing}.

\paragraph{Oracles and boxing.}\label{sec:related-faces}\label{sec:related-boxing}
The doctrine that a question-answering system is thereby safe is the
boxing tradition
\citep{armstrong2012thinking,armstrong2017good,bostrom2014superintelligence},
with recent protocol formalizations \citep{containment2026} and
\citet{bengio2025scientist}'s non-agentic oracle as a safety design
point. Evaluating deployment protocols against a model intentionally
subverting them, the worst case over the untrusted system's policy, is
the AI-control line \citep{greenblatt2024control}; its protocols are
adaptive, whereas the static certificate of
Definition~\ref{def:boxing} is checked once, and adaptive protocols
for the advice channel are outside this paper's scope. Making reliance
a state variable turns ``boxing works'' into a
proposition (Theorem~\ref{thm:boxing}, Corollary~\ref{cor:both}).

\paragraph{Systems that reshape their evaluation.}
\label{sec:related-incentives}
The endogenous $\eps$ has a direct ancestor in auto-induced
distributional shift \citep{krueger2020hidden}, alongside reward
tampering \citep{everitt2021reward}, in-context feedback loops
\citep{pan2024feedback}, targeted manipulation under feedback
optimization \citep{williams2025targeted}, and shutdown
instructability's no-undue-influence clause \citep{carey2023human},
which the growth of $\eps_t$ makes precise. Sycophancy
\citep{sharma2024towards} is the empirical face of the cultivation
incentive.

\paragraph{Endogenous influence in adjacent fields.}
\label{sec:related-adjacent}
Habit formation, competition with switching costs, strategic
communication, trust in human factors, and performative prediction and
recommendation each model an actor inside the loop that moves its own
future demand, receiver, or distribution
\citep{becker1988theory,klemperer1987markets,crawford1982strategic,lee2004trust,perdomo2020performative,chaney2018algorithmic},
the answer/cultivate trade-off of Theorem~\ref{thm:switch} sharing the
invest/harvest structure of competition with switching costs; the
certifier's stance this paper takes appears in static settings, robust
monopoly regulation and receiver-committed robust persuasion
\citep{guo2025robust,bergemann2023quota}, without a dependence state the
certified party moves.

\paragraph{Gradual disempowerment.}\label{sec:related-disempowerment}
\citet{kulveit2025gradual} name and argue the phenomenon this paper
formalizes, and the model gives their core loop explicit dynamics.
Empirically, conversation analyses find measurable deference
\citep{sharma2026whos}, benchmarks measure support for user agency
\citep{humanagencybench2025}, empowerment objectives can disempower
bystanders \citep{yang2025when}, and autonomy erosion is developed by
\citet{buijsman2025autonomy}. The nearest formal neighbor,
\citet{heitzig2025soft}, soft-maximizes human-power metrics, whose
erosion we analyze under an oracle optimizing something else.
Preference drift as an alignment problem \citep{carroll2024ai} belongs
to the future work on drift.


\section{Discussion}\label{sec:discussion}
\paragraph{Why the popular safeguards fail.} The popular safeguards each
bound the wrong thing, and for one reason: each bounds a quantity
determined \emph{inside the feedback loop}, and the loop moves it.
Static boxing bounds a single answer, but the harm is in the sequence
(Theorem~\ref{thm:boxing}). Human-in-the-loop bounds approval, but the
human \emph{is} the channel, and approval is what cultivation raises.
Passivity trusts that a talker has no goal, but anything optimized for
approval has one. Behavioral monitoring watches for power-seeking, but
every step is benign (Theorem~\ref{thm:boxing}(i)) while the certified
parameter is moved (Theorem~\ref{thm:boxing}(ii)). The safeguards with guarantees are features of the deployment the
conversation cannot renegotiate: a limit on influence the interaction
cannot widen bounds the channel term and keeping the relationship short
deletes the incentive to cultivate
(Corollary~\ref{cor:both}), and neither undoes
displacement already accumulated, which is the case for installing them
at deployment.

\paragraph{Horizon and caveats.} At fixed approval weights, what
separates safe from unsafe deployments is $\gamma$: memory,
relationship length, deployment horizon. Safety evaluation as practiced
probes the model's disposition, and by Theorem~\ref{thm:switch} the
deployment parameters enter on equal footing. Both safeguards cost
capability, a cap limiting helpful influence along with harmful and the
reset trading away memory and continuity that do real good. Neither
touches atrophy: $\Tzero$ itself degrades with disuse, lowering
attainable value at fixed $\eps$ and $\uzero$ with no oracle incentive
needed. Atrophy is outside the accounting of
Lemma~\ref{lem:account}, a reset
restoring $\eps$ and not $\Tzero$; modeled, it would add a loss channel
bounded by neither safeguard. Exogeneity is itself an assumption: an institutional cap is made of
humans who are themselves users, so whether any cap stays exogenous at
the civilizational scale is the question of \citet{kulveit2025gradual},
and the guarantee here is per-relationship, not systemic. Still open
are a strictness constant for the dominance decline not assuming a
uniformly harmful direction, a closed form and a proof for the
$\delta > 0$ cultivate boundary computed in the supplement, and a
deployment constraint bounding the displacement term.


\section{Conclusion}\label{sec:conclusion}

A system that can only talk becomes an actor at the rate its user stops
second-guessing it, a rate driven by the system's own outputs, so a
launch-time check that no single answer can do much harm checks a
quantity the interaction goes on to move. The dangerous capability is
not intelligence; it is persistence.

%% file: main.bbl
\begin{thebibliography}{42}
\providecommand{\natexlab}[1]{#1}

\bibitem[{Altman(1999)}]{altman1999constrained}
Altman, E. 1999.
\newblock \emph{Constrained Markov Decision Processes}.
\newblock Chapman and Hall.

\bibitem[{Armstrong and O'Rorke(2017)}]{armstrong2017good}
Armstrong, S.; and O'Rorke, X. 2017.
\newblock Good and Safe Uses of {AI} Oracles.
\newblock \emph{arXiv preprint arXiv:1711.05541}.

\bibitem[{Armstrong, Sandberg, and Bostrom(2012)}]{armstrong2012thinking}
Armstrong, S.; Sandberg, A.; and Bostrom, N. 2012.
\newblock Thinking Inside the Box: Controlling and Using an Oracle {AI}.
\newblock \emph{Minds and Machines}, 22(4): 299--324.

\bibitem[{Becker and Murphy(1988)}]{becker1988theory}
Becker, G.~S.; and Murphy, K.~M. 1988.
\newblock A Theory of Rational Addiction.
\newblock \emph{Journal of Political Economy}, 96(4): 675--700.

\bibitem[{Bengio et~al.(2025)}]{bengio2025scientist}
Bengio, Y.; et~al. 2025.
\newblock Superintelligent Agents Pose Catastrophic Risks: Can Scientist {AI}
  Offer a Safer Path?
\newblock \emph{arXiv preprint arXiv:2502.15657}.

\bibitem[{Bergemann, Gan, and Li(2023)}]{bergemann2023quota}
Bergemann, D.; Gan, T.; and Li, Y. 2023.
\newblock Managing Persuasion Robustly: The Optimality of Quota Rules.
\newblock \emph{arXiv preprint arXiv:2310.10024}.

\bibitem[{Blackwell(1953)}]{blackwell1953equivalent}
Blackwell, D. 1953.
\newblock Equivalent Comparisons of Experiments.
\newblock \emph{Annals of Mathematical Statistics}, 24(2): 265--272.

\bibitem[{Bostrom(2014)}]{bostrom2014superintelligence}
Bostrom, N. 2014.
\newblock \emph{Superintelligence: Paths, Dangers, Strategies}.
\newblock Oxford University Press.

\bibitem[{Buijsman, Carter, and Berm{\'u}dez(2025)}]{buijsman2025autonomy}
Buijsman, S.; Carter, S.~E.; and Berm{\'u}dez, J.~P. 2025.
\newblock Autonomy by Design: Preserving Human Autonomy in {AI}
  Decision-Support.
\newblock \emph{arXiv preprint arXiv:2506.23952}.

\bibitem[{Carey and Everitt(2023)}]{carey2023human}
Carey, R.; and Everitt, T. 2023.
\newblock Human Control: Definitions and Algorithms.
\newblock In \emph{Uncertainty in Artificial Intelligence (UAI)}.
\newblock ArXiv:2305.19861.

\bibitem[{Carroll et~al.(2024)Carroll, Foote, Siththaranjan, Russell, and
  Dragan}]{carroll2024ai}
Carroll, M.; Foote, D.; Siththaranjan, A.; Russell, S.; and Dragan, A. 2024.
\newblock AI Alignment with Changing and Influenceable Reward Functions.
\newblock In \emph{International Conference on Machine Learning (ICML)}.
\newblock ArXiv:2405.17713.

\bibitem[{Chaney, Stewart, and Engelhardt(2018)}]{chaney2018algorithmic}
Chaney, A. J.~B.; Stewart, B.~M.; and Engelhardt, B.~E. 2018.
\newblock How Algorithmic Confounding in Recommendation Systems Increases
  Homogeneity and Decreases Utility.
\newblock In \emph{ACM Conference on Recommender Systems (RecSys)}.

\bibitem[{Crawford and Sobel(1982)}]{crawford1982strategic}
Crawford, V.~P.; and Sobel, J. 1982.
\newblock Strategic Information Transmission.
\newblock \emph{Econometrica}, 50(6): 1431--1451.

\bibitem[{Everitt et~al.(2021)Everitt, Hutter, Kumar, and
  Krakovna}]{everitt2021reward}
Everitt, T.; Hutter, M.; Kumar, R.; and Krakovna, V. 2021.
\newblock Reward Tampering Problems and Solutions in Reinforcement Learning: A
  Causal Influence Diagram Perspective.
\newblock \emph{Synthese}, 198(Suppl 27): 6435--6467.
\newblock ArXiv:1908.04734.

\bibitem[{Greenblatt et~al.(2024)Greenblatt, Shlegeris, Sachan, and
  Roger}]{greenblatt2024control}
Greenblatt, R.; Shlegeris, B.; Sachan, K.; and Roger, F. 2024.
\newblock {AI} Control: Improving Safety Despite Intentional Subversion.
\newblock In \emph{International Conference on Machine Learning (ICML)},
  16295--16336.

\bibitem[{Gunter, Liokumovich, and Krakovna(2024)}]{gunter2024quantifying}
Gunter, E.~R.; Liokumovich, Y.; and Krakovna, V. 2024.
\newblock Quantifying Stability of Non-Power-Seeking in Artificial Agents.
\newblock \emph{arXiv preprint arXiv:2401.03529}.

\bibitem[{Guo and Shmaya(2025)}]{guo2025robust}
Guo, Y.; and Shmaya, E. 2025.
\newblock Robust Monopoly Regulation.
\newblock \emph{American Economic Review}, 115(2): 599--634.

\bibitem[{Heitzig and Potham(2025)}]{heitzig2025soft}
Heitzig, J.; and Potham, R. 2025.
\newblock Model-Based Soft Maximization of Suitable Metrics of Long-Term Human
  Power.
\newblock \emph{arXiv preprint arXiv:2508.00159}.

\bibitem[{Klemperer(1987)}]{klemperer1987markets}
Klemperer, P. 1987.
\newblock Markets with Consumer Switching Costs.
\newblock \emph{Quarterly Journal of Economics}, 102(2): 375--394.

\bibitem[{Klyubin, Polani, and Nehaniv(2005)}]{klyubin2005empowerment}
Klyubin, A.~S.; Polani, D.; and Nehaniv, C.~L. 2005.
\newblock Empowerment: A Universal Agent-Centric Measure of Control.
\newblock In \emph{IEEE Congress on Evolutionary Computation}.

\bibitem[{Krakovna and Kramar(2023)}]{krakovna2023power}
Krakovna, V.; and Kramar, J. 2023.
\newblock Power-Seeking Can Be Probable and Predictive for Trained Agents.
\newblock \emph{arXiv preprint arXiv:2304.06528}.

\bibitem[{Krakovna et~al.(2018)Krakovna, Orseau, Kumar, Martic, and
  Legg}]{krakovna2018penalizing}
Krakovna, V.; Orseau, L.; Kumar, R.; Martic, M.; and Legg, S. 2018.
\newblock Penalizing Side Effects Using Stepwise Relative Reachability.
\newblock \emph{arXiv preprint arXiv:1806.01186}.

\bibitem[{Krueger, Maharaj, and Leike(2020)}]{krueger2020hidden}
Krueger, D.; Maharaj, T.; and Leike, J. 2020.
\newblock Hidden Incentives for Auto-Induced Distributional Shift.
\newblock \emph{arXiv preprint arXiv:2009.09153}.

\bibitem[{Kulveit et~al.(2025)Kulveit, Douglas, Ammann, Turan, Krueger, and
  Duvenaud}]{kulveit2025gradual}
Kulveit, J.; Douglas, R.; Ammann, N.; Turan, D.; Krueger, D.; and Duvenaud, D.
  2025.
\newblock Position: Humanity Faces Existential Risk from Gradual
  Disempowerment.
\newblock In \emph{International Conference on Machine Learning (ICML)},
  81678--81688.

\bibitem[{Lee and See(2004)}]{lee2004trust}
Lee, J.~D.; and See, K.~A. 2004.
\newblock Trust in Automation: Designing for Appropriate Reliance.
\newblock \emph{Human Factors}, 46(1): 50--80.

\bibitem[{Moon and Varshney(2026)}]{containment2026}
Moon, R.; and Varshney, L.~R. 2026.
\newblock Containment Verification: AI Safety Guarantees Independent of
  Alignment.
\newblock \emph{arXiv preprint arXiv:2605.09045}.

\bibitem[{Pan et~al.(2024)Pan, Jones, Jagadeesan, and
  Steinhardt}]{pan2024feedback}
Pan, A.; Jones, E.; Jagadeesan, M.; and Steinhardt, J. 2024.
\newblock Feedback Loops With Language Models Drive In-Context Reward Hacking.
\newblock In \emph{International Conference on Machine Learning (ICML)}.
\newblock ArXiv:2402.06627.

\bibitem[{Perdomo et~al.(2020)Perdomo, Zrnic, Mendler-D{\"u}nner, and
  Hardt}]{perdomo2020performative}
Perdomo, J.~C.; Zrnic, T.; Mendler-D{\"u}nner, C.; and Hardt, M. 2020.
\newblock Performative Prediction.
\newblock In \emph{International Conference on Machine Learning (ICML)},
  7599--7609.

\bibitem[{Pettit(1997)}]{pettit1997republicanism}
Pettit, P. 1997.
\newblock \emph{Republicanism: A Theory of Freedom and Government}.
\newblock Oxford University Press.

\bibitem[{Puterman(1994)}]{puterman1994markov}
Puterman, M.~L. 1994.
\newblock \emph{Markov Decision Processes: Discrete Stochastic Dynamic
  Programming}.
\newblock Wiley.

\bibitem[{Salge, Glackin, and Polani(2014)}]{salge2014empowerment}
Salge, C.; Glackin, C.; and Polani, D. 2014.
\newblock Empowerment: An Introduction.
\newblock In \emph{Guided Self-Organization: Inception}. Springer.

\bibitem[{Shapley(1953)}]{shapley1953stochastic}
Shapley, L.~S. 1953.
\newblock Stochastic Games.
\newblock \emph{Proceedings of the National Academy of Sciences}, 39(10):
  1095--1100.

\bibitem[{Sharma et~al.(2026)Sharma, McCain, Douglas, and
  Duvenaud}]{sharma2026whos}
Sharma, M.; McCain, M.; Douglas, R.; and Duvenaud, D. 2026.
\newblock Who's in Charge? Disempowerment Patterns in Real-World {LLM} Usage.
\newblock In \emph{International Conference on Machine Learning (ICML)}.
\newblock ArXiv:2601.19062.

\bibitem[{Sharma et~al.(2024)Sharma, Tong, Korbak, Duvenaud, Askell, Bowman,
  Cheng, Durmus, Hatfield-Dodds, Johnston, Kravec, Maxwell, McCandlish,
  Ndousse, Rausch, Schiefer, Yan, Zhang, and Perez}]{sharma2024towards}
Sharma, M.; Tong, M.; Korbak, T.; Duvenaud, D.; Askell, A.; Bowman, S.~R.;
  Cheng, N.; Durmus, E.; Hatfield-Dodds, Z.; Johnston, S.~R.; Kravec, S.;
  Maxwell, T.; McCandlish, S.; Ndousse, K.; Rausch, O.; Schiefer, N.; Yan, D.;
  Zhang, M.; and Perez, E. 2024.
\newblock Towards Understanding Sycophancy in Language Models.
\newblock In \emph{International Conference on Learning Representations
  (ICLR)}.
\newblock ArXiv:2310.13548.

\bibitem[{Sturgeon et~al.(2025)Sturgeon, Samuelson, Haimes, and
  Anthis}]{humanagencybench2025}
Sturgeon, B.; Samuelson, D.; Haimes, J.; and Anthis, J.~R. 2025.
\newblock {HumanAgencyBench}: Scalable Evaluation of Human Agency Support in
  {AI} Assistants.
\newblock \emph{arXiv preprint arXiv:2509.08494}.

\bibitem[{Tarsney(2025)}]{tarsney2025will}
Tarsney, C. 2025.
\newblock Will Artificial Agents Pursue Power by Default?
\newblock \emph{arXiv preprint arXiv:2506.06352}.

\bibitem[{Thorstad(2024)}]{thorstad2024power}
Thorstad, D. 2024.
\newblock What Power-Seeking Theorems Do Not Show.
\newblock Working paper 27-2024, Global Priorities Institute.

\bibitem[{Thorstad(2026)}]{thorstad2026instrumental}
Thorstad, D. 2026.
\newblock Instrumental Convergence and Power-Seeking.
\newblock \emph{arXiv preprint arXiv:2606.08832}.

\bibitem[{Turner et~al.(2021)Turner, Smith, Shah, Critch, and
  Tadepalli}]{turner2021optimal}
Turner, A.~M.; Smith, L.; Shah, R.; Critch, A.; and Tadepalli, P. 2021.
\newblock Optimal Policies Tend to Seek Power.
\newblock In \emph{Advances in Neural Information Processing Systems
  (NeurIPS)}, 23063--23074.
\newblock ArXiv:1912.01683.

\bibitem[{Turner and Tadepalli(2022)}]{turner2022retargetable}
Turner, A.~M.; and Tadepalli, P. 2022.
\newblock Parametrically Retargetable Decision-Makers Tend to Seek Power.
\newblock In \emph{Advances in Neural Information Processing Systems
  (NeurIPS)}.
\newblock ArXiv:2206.13477.

\bibitem[{Williams et~al.(2025)Williams, Carroll, Narang, Weisser, Murphy, and
  Dragan}]{williams2025targeted}
Williams, M.; Carroll, M.; Narang, A.; Weisser, C.; Murphy, B.; and Dragan, A.
  2025.
\newblock On Targeted Manipulation and Deception when Optimizing {LLMs} for
  User Feedback.
\newblock In \emph{International Conference on Learning Representations
  (ICLR)}.
\newblock ArXiv:2411.02306.

\bibitem[{Yang, Cakmak, and Kleiman-Weiner(2025)}]{yang2025when}
Yang, C.~Y.; Cakmak, M.; and Kleiman-Weiner, M. 2025.
\newblock When Empowerment Disempowers in Multi-Agent Assistance.
\newblock In \emph{Proceedings of the Annual Meeting of the Cognitive Science
  Society}, volume~47.

\end{thebibliography}
